\documentclass[letterpaper, 10 pt, conference]{ieeeconf}  

\IEEEoverridecommandlockouts 
\usepackage[noadjust,space]{cite}
\usepackage{svg}
\usepackage{amsmath,amssymb,amsfonts}
\usepackage{algorithmic}
\usepackage{graphicx}
\usepackage{textcomp}
\usepackage{xcolor}
\usepackage{tabularx}
\usepackage{dsfont}
\usepackage[]{subcaption}
\usepackage{colortbl} 
\usepackage{hhline} 
\usepackage[colorlinks=true,allcolors=.,urlcolor=blue,]{hyperref}
\usepackage{multicol}
\usepackage{flushend}
\usepackage{multirow}
\usepackage{float}

\def\tablename{Table}
\DeclareCaptionLabelFormat{mytablestyle}{\tablename~\thetable}
\renewcommand{\thetable}{\arabic{table}}

\DeclareMathOperator*{\argmax}{\arg\!\max}

\graphicspath{ {./images/} }

\makeatletter
\def\convertto#1#2{\strip@pt\dimexpr #2*65536/\number\dimexpr 1#1}
\makeatother

\title{\LARGE \bf
	ProcNet: Deep Predictive Coding Model for Robust-to-occlusion Visual Segmentation and Pose Estimation
}

\author{Michael Zechmair$^{1}$, Alban Bornet$^{2}$,  and Yannick Morel$^{1}$
	\thanks{$^{1}$Michael Zechmair and Yannick Morel are with Faculty of Psychology and Neuroscience, Maastricht University, %
		Maastricht, The Netherlands, %
		{\tt\small \{m.zechmair,y.morel\}@unimaas.nl}}%
	\thanks{$^{2}$Alban Bornet is with with Department of Radiology and Medical Informatics, University of Geneva, Geneva, Switzerland, {\tt \small alban.bornet@unige.ch} and was with Laboratory of Psychophysics, Brain Mind Institute, EPFL, Lausanne, Switzerland}%
	\thanks{All code and datasets have been made publicly available \href{https://github.com/Mike2208/prednet\_localization\_results}{here}}
	\thanks{This research has received funding from the European Union's Horizon 2020 Framework Programme for Research and Innovation under the Specific Grant Agreement No. 785907 (Human Brain Project SGA3).}%
}

\begin{document}
	\newcolumntype{a}{>{\columncolor{lightgray}}c}
	
	\maketitle
	\begin{abstract}
	Systems involving human-robot collaboration necessarily require that steps be taken to ensure safety of the participating human. This is usually achievable if accurate, reliable estimates of the human's pose are available. In this paper, we present a deep Predictive Coding (PC) model supporting visual segmentation, which we extend to pursue pose estimation. The model is designed to offer robustness to the type of transient occlusion naturally occurring when human and robot are operating in close proximity to one another. Impact on performance of relevant model parameters is assessed, and comparison to an alternate pose estimation model (NVIDIA's PoseCNN) illustrates efficacy of the proposed approach.
	\end{abstract}
	
	\section{Introduction} %
Over the past several decades, robotic technology has become prevalent in industrial processing and manufacturing lines. Robots have been used to pursue task automation and reduce human workload. Robotic systems are particularly well-suited to performing repeatable, structured processes. Recent research has focused on expanding the capabilities of robotic technology to develop systems displaying a greater degree of autonomy, and the ability of adapting to environmental, possibly task-altering factors (\cite{schuh2017}). Advances in the area have permitted the automation of an increasingly broader range of processes. Key to enabling automation of progressively less structured processes is the ability to reliably perceive the system's surrounding, in particular the relative configuration of objects of interest, which the system is required to interact with in a controlled, predetermined manner to perform its designated task. This is of particular import in situations of expected human-robot collaboration, in which ensuring human safety necessarily requires that the robot be able to ascertain space occupancy of the human in proximity at all times. Exploiting such information, the robotic system is typically able to adjust its movements to avoid collisions, but also assess human intentions, and pursue safe, effective collaboration.

Existing approaches to object localization or pose estimation commonly rely on visual modalities. The visual perception process typically involves the identification of spatial (or temporal) patterns within streams of frames captured by camera and, based on identified features, inferring information pertaining to objects of interest in the scene. A broad range of modern computer vision models rely on Convolutional Neural Networks (CNNs) to address visual segmentation tasks (\cite{chen2018encoder, girshick2014rich, he2017mask, long2015fully, ronneberger2015u}). Their versatility allows CNNs to achieve excellent performance for a large number of other vision-related tasks, including image recognition (\cite{krizhevsky2012imagenet, tan2019efficientnet}), image generation (\cite{goodfellow2014generative, karras2019style, radford2015unsupervised}), and scene rendering (\cite{eslami2018neural, xiao2018unified}). CNNs have proven useful for a meaningful number of image-based practical applications such as medical imaging (\cite{shen2015multi, shin2016deep}), autonomous driving (\cite{xu2017end}), or manufacturing (\cite{sateesh2016deep, wen2020transfer}). However, CNNs suffer from well-established limitations in the presence of visual occlusion (\cite{fawzi2016measuring, kortylewski2020combining, mcallister2017concrete}). For visual segmentation for instance, segmentation masks typically fail to reflect the occluded object's shape, and the occluding object's outline may be mistaken for the boundaries of the occluded object. The problem is significant in collaborative robotic use-cases, where humans and robots work in close proximity (\cite{zhu2022challenges}). Specifically, situations in which human and robotic arm may share a common workspace lead to substantial and frequent occlusion, impairing inferences made by CNN models.

Human visual processing still dramatically outperforms modern machine vision. In particular, human vision has demonstrated a remarkable degree of robustness to occlusion (\cite{rajaei2019beyond, tang2018recurrent, wyatte2012limits}). Affording consideration to the manner in which human visual processing manages to mitigate the impact of occlusion may yield insights allowing to improve CNNs’ robustness. More broadly, careful consideration of the different neural mechanism involved in human vision, such as those intervening within the human visual cortex, may prove of benefit in the perspective of improving the flexibility and generalization capabilities of visual neural models. When pursuing brain-inspired approaches however, one should remain mindful of that fact that, although CNNs approach human-like performance in several complex visual tasks and provide the best models of image-evoked population response in the primate visual cortex (\cite{schrimpf2018brain, yamins2014performance}), there exits fundamental differences between the type of processing involved in human vision and that implemented by CNNs. In other words, while CNNs may achieve human-like performance, that does not necessarily imply they implement human-like computations. For example, although the specific role of feedback connections in the human visual cortex remains a matter of debate (\cite{olshausen2006other, van2020going}), there exists a broad consensus on their functional significance. However, typical CNNs lack such pathways, largely remaining strictly feedforward models. Achieving a better understanding of the significance of such discrepancies is likely to prove beneficial to developing more human-like models, reflecting human qualities of robustness that have eluded CNNs. In this perspective, there exists a large corpus of results from vision psychophysics paradigms that CNNs cannot explain. For example, visual crowding experiments have shown that the human visual cortex integrates information across large portions of the visual field (\cite{malania2007grouping, vickery2009supercrowding}). These experiments suggest that, in human vision, high-level context about the global configuration of the visual input strongly affects local and low-level information processing (\cite{manassi2016crowding}). In contrast, CNNs prove unable to reproduce such results as they typically rely on feedforward and local operations exclusively (\cite{bornet2021global}). Identifying what is missing from CNNs to account for global aspects of crowding may prove helpful in understanding the manner in which they differ from human visual processing. Recent studies have showed that the only models of human vision able to explain the global aspects of visual crowding include explicit recurrent grouping and segmentation processes (\cite{bornet2021shrinking, doerig2019beyond}). For example, adding dynamic routing to CNNs (capsule networks) (\cite{sabour2017dynamic}) or illusory contour mechanisms (\cite{francis2017neural}), both of which instantiate grouping and segmentation, allows to better match human behaviour in visual crowding paradigms (\cite{bornet2021shrinking, doerig2020capsule}). This line of evidence suggests that one computational function of recurrent processing in the human visual cortex is to efficiently discriminate between features to integrate ({\it grouping}) and features to segregate ({\it segmentation}). These processes help the brain cope with complex input data (featuring occlusion, reflections, noise, etc.) and refine low-level information based on high-level context. In computer vision for instance, adding feedback processing to CNNs has been shown to prove helpful when performing inferences from partial information (\cite{rajaei2019beyond, tang2018recurrent}). A computational paradigm that allows to reflect key features in human vision is Predictive Coding (PC, \cite{millidge2021predictive}), its use to support visual processing is thus of particular interest.

In this paper, we build upon psycho-physical insights to create a visual segmentation model providing a degree of robustness to occlusion. More specifically, we present a method able to learn to produce segmentation masks based on Predictive Coding, and infer likeliest pose by comparing the  segmentation mask produced by the PC model to a range of candidate masks, reflecting different candidate poses. Gradient following is used to support convergence towards the pose whose mask provides the best fit to the PC-produced mask. The main contribution of the paper consists in proposing a novel approach to visual segmentation, which offers qualities of robustness to transient visual occlusion: ProcNet. This paper is organized as follows. Section II describes the visual segmentation and pose estimation model. Results of numerical simulations are presented in Section III. Section IV concludes this paper.
	
	\section{Pose Estimation Algorithm} %
Predictive Coding paradigms are based on the notion that the brain's core functions revolve around the minimization of prediction errors computed by neural circuitry (\cite{millidge2021predictive}). A meaningful number of robotic systems, attempting to emulate some of the brain's functions or qualities, have come to rely on algorithms developed using this paradigm. As previously discussed, one such area of research is perception, where the ability to approach human object detection, scene understanding, or pose estimation capabilities would be desirable. In the following, we present a vision-based method able to learn to both determine an object's shape and estimate its pose, in the face of transient, partial occlusion.
	
	\subsection{Predictive Coding-based vision segmentation} %
	\begin{figure}
	\vspace{0.5em}
	\centering
	\def\svgwidth{\columnwidth}
	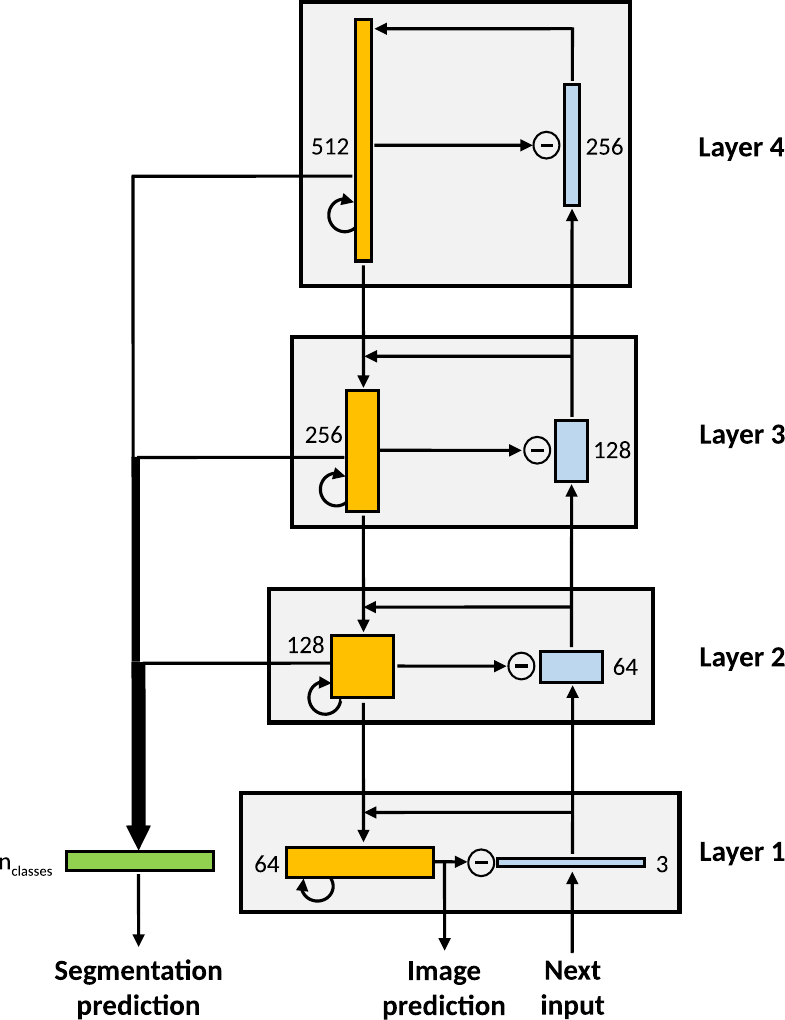
	\caption{Overview of the ProcNet architecture, derived from PredNet. Width of each coloured box indicates the spatial scale of the layer component (the larger, the more resolution), height indicates the number of feature maps.}
	\label{fig:prednet_model_diag_a}
\end{figure}

\begin{figure}
	\vspace{0.7em}
	\centering
	\def\svgwidth{\columnwidth}
	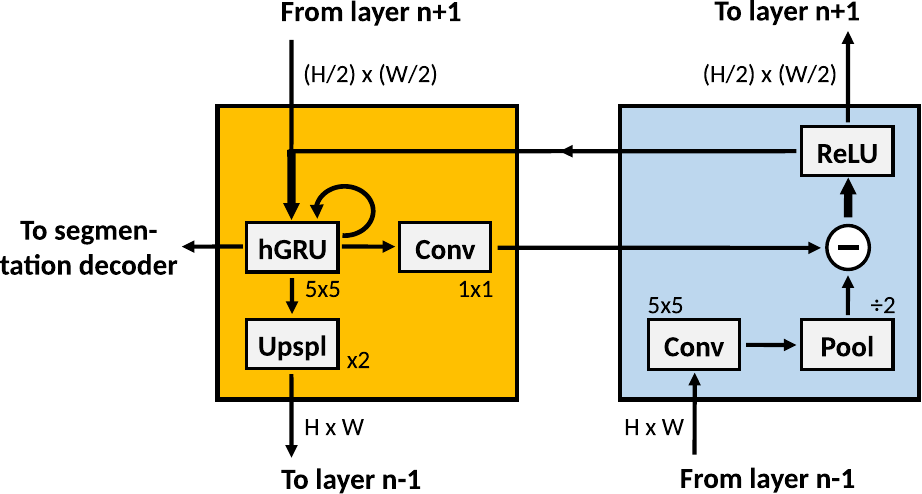
	\caption{Detailed view of computations performed at each ProcNet layer. Numbers represent kernel size for hGRU and convolution (Conv) operations, scaling factor for the up-sampling (Upspl) and Pool operations. Note that Conv and Pool operations are absent in the first layer of ProcNet.}
	\label{fig:prednet_model_diag_b}
\end{figure}

\begin{figure}
	\vspace{0.5em}
	\centering
	\def\svgwidth{\columnwidth}
	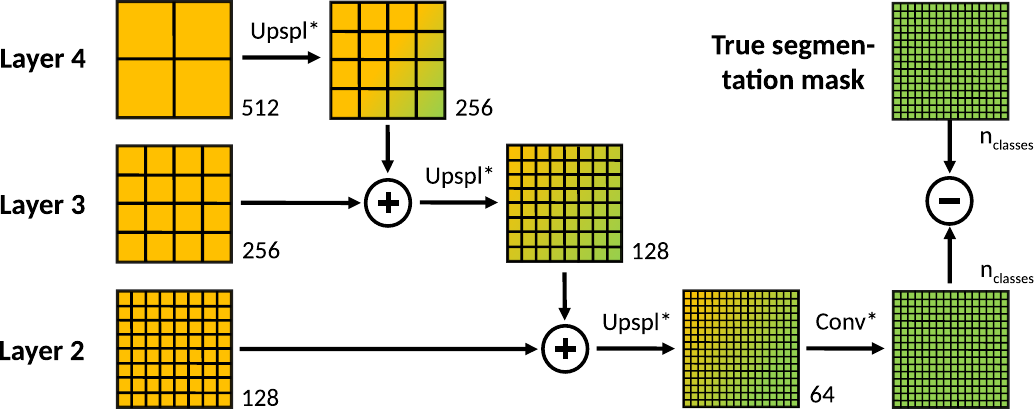
	\caption{Detailed view describing the manner in which latent activity of the decoded layers' representation components are combined in the segmentation prediction process.}
	\vspace{-1em}
	\label{fig:prednet_model_diag_c}
\end{figure}

Visual segmentation entails the estimation of an object's outline from a given camera frame. It requires determining the portion of the camera frame that belongs to the considered object. We use an approach derived from deep Predictive Coding (as described in \cite{lotter2016deep}) to generate the segmentation mask that delineates the object's shape in the image. 
In particular, this mask is generated by a neural model based on the PredNet architecture (see Fig. \ref{fig:prednet_model_diag_a}, \ref{fig:prednet_model_diag_b}, and \ref{fig:prednet_model_diag_c}). PredNet is composed of a stack of layers that represent information and compute prediction errors at that particular layer's level. Each layer is composed of a representation component (yellow boxes) and a prediction error computation component (blue boxes). Representation components use top-down information (arrows pointing downwards) and the error signal computed by the prediction error component (arrows pointing left) to generate an accurate prediction of the incoming bottom-up input of the considered layer. They include recurrent units (recursive arrows), to allow retention of information over time. In the downstream direction, max pooling operations decrease spatial resolution, and convolutions compute local feature maps. The prediction error computation components subtract the prediction generated by the representation component at that layer (arrows pointing right), from the input received from the upstream layer (arrows pointing upwards). Their output constitutes the local prediction error signal used in the layer. 

Streams of camera frames are presented to the first layer of the network, and, for each presented frame, a prediction error signal is computed and propagated to the downstream layer, while top-down activity is projected from the higher layer, based on the activity of the network related to previous frames. Similar interactions across top-down and bottom-up signals occurs at each layer. At every time step, the sum of all layers’ prediction error is computed and constitutes the self-supervised loss of the network. 

A detailed view of the computations performed within each layer of ProcNet is provided in Fig. \ref{fig:prednet_model_diag_b}. In the prediction error computation component (blue box), the output of the upstream layer is sent to a convolution (kernel size 5) and a max-pooling operation (scale factor 2). Then, the result produced by these operations is subtracted from the prediction generated by the representation component ({\it prediction error}; minus sign). Both positive and negative prediction errors (i.e. prediction subtracted from input and input subtracted from prediction) are concatenated along the feature dimension (double-lined arrow pointing upwards). Finally, the concatenated prediction error undergoes a ReLU operation and serves as input for the downstream layer. The produced signal constitutes the {\it prediction error signal} of each layer.

In the representation component (yellow box), the output of the downstream layer is concatenated along the feature map dimension with the prediction error of the same layer (double-lined arrow pointing downwards). We replaced the LSTM units of the original implementation of PredNet with horizontal Gated Recurrent Units (hGRU). hGRUs are specifically designed to integrate spatial information over time and have been empirically shown to exhibit illusory contours. Our assumption is that the ability of hGRUs to capture illusory contours contributes to more accurate reconstruction of segmentation masks, especially for partially occluded objects. The underlying computations performed by hGRU cells are described in \cite{doerig2020capsule}. In particular, hGRU cells implement a local excitatory-inhibitory loop between two hidden layers linked by convolution operations (kernel size 5). This loop can learn temporal dependencies between visual elements by spreading information locally. Therefore, our implementation can account for information from previous time steps when determining the current time instant's segmentation masks. For moving objects, this additional information (corresponding to the object's past shape configuration) can be extrapolated to help determine the current mask, even in a situation in which the current frame only provides incomplete information (e.g. if view of the object is occluded). A convolution layer (kernel size 1; used to adjust the number of feature maps) generates the prediction that is sent to the prediction error computation component (arrow going from the yellow to the blue box). This output is presented to an up-sampling operator (scale factor 2), the output of which serves as the top-down input for the upstream layer. Finally, the output of hGRU cells is sent to the segmentation decoder module (arrow pointing left). The decoding module (in Fig. \ref{fig:prednet_model_diag_c}) relies on the output of the representation component of all layers except the first one (which is used for image prediction). The difference between the decoded segmentation mask and the true segmentation mask constitutes the supervised loss of the network. It is computed as a weighed average of Dice (\cite{sudre2017generalised}) and Focal scores (\cite{lin2017focal}) between the segmentation decoder module's output and the ground truth. An array of pixels within which generated segmentation masks are represented is noted $m \in \mathds{N}^{h \times w}$, with $h$, $w \in \mathds{N}$ describing the height and width of the camera image in pixels, respectively. The values in $m$ correspond to the index of a detected object at the individual pixel coordinates, with a value of $0$ indicating no detected object.

\begin{figure}
	\centering
	\includegraphics[width=\columnwidth]{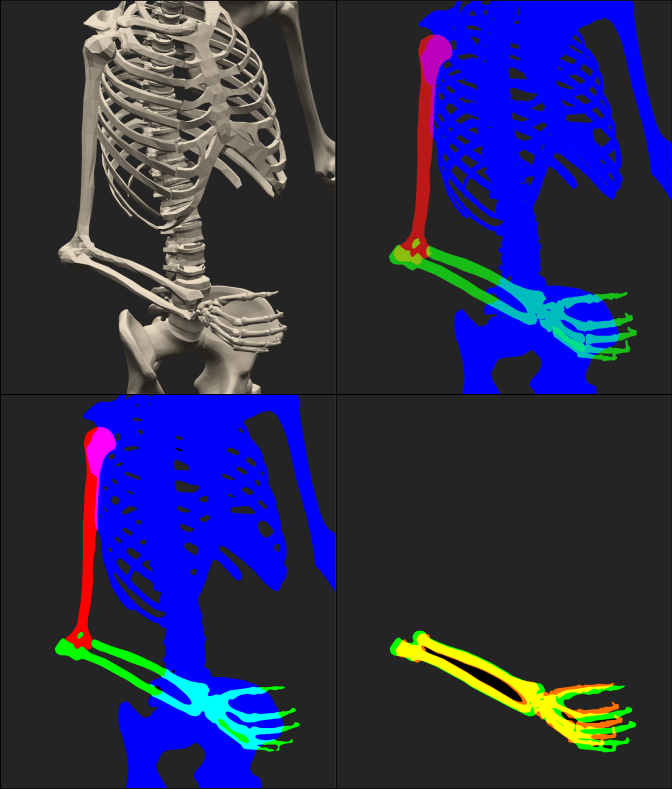}
	\caption{Segmentation and pose estimation of skeletal human forearm. The top-left image shows the camera frame, top-right and bottom-left show the ground truth and the ProcNet-generated segmentation masks, respectively. Different colors denote distinct, segmented parts of the skeleton. Bottom right shows an example of pose estimation, where the segmentation mask associated with the estimated pose overlaps with the mask produced by the predictive coding model.}
	\label{fig:vis_segmentation}
	\vspace{-0.5em}
\end{figure}

\begin{figure}
	\begin{center}
		\includegraphics[width=0.95\columnwidth]{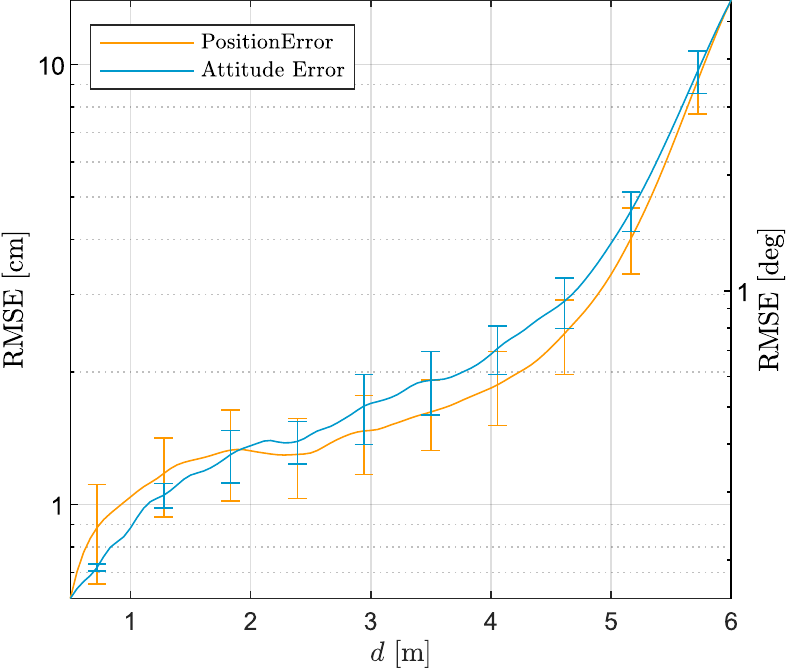}%
		\vspace{1em}
		\includegraphics[width=1\columnwidth]{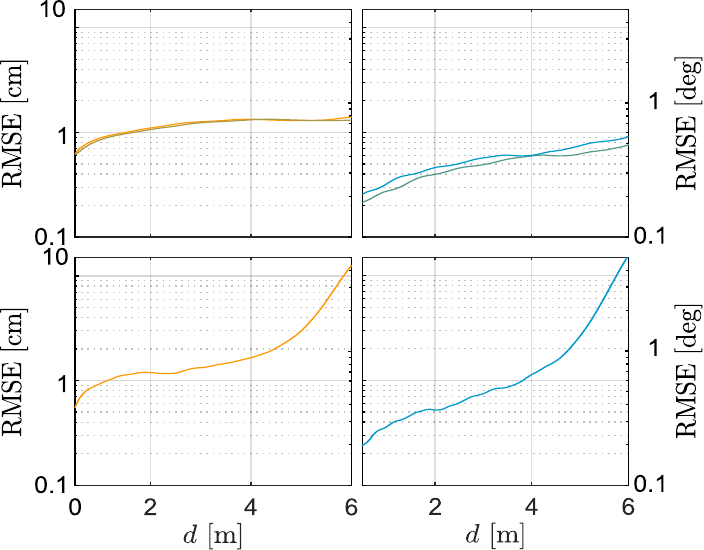}
	\end{center}
	\caption{Position and attitude estimation error represented as a function of distance $d$ from the camera (top); position errors in the vertical plane (top left), attitude errors in yaw and pitch (top right), position error in the depth direction (bottom left), roll error (bottom right).}
	\label{fig:vis_sensor_dist_err}
\end{figure}

	\subsection{Object Pose Estimation} %
	\begin{table}
	\centering
	\begin{tabular}{|c|c|c|c||c|c|c|}
		\hline 
		\textbf{\shortstack{ProcNet \\ Layers}} & \shortstack{\textbf{Act.} \\ \textbf{Fcn.}} & \textbf{\shortstack{Axonal \\ Delay}} & \textbf{\shortstack{Pred. \\ Loss}} & \textbf{\shortstack{Dice \\ Loss}} & \textbf{\shortstack{Focal \\ Loss}} & \textbf{\shortstack{Avg. \\ Loss}}\\
		\hline \hline
		\multirow{12}*{\textbf{3}} & \multirow{4}*{Conv} & 0 & 0 & 0.4972 & 0.5143 & 0.5058 \\
		& & 0 & 1  & 0.3943 & 0.4073 & 0.4008 \\
		& & 1  & 0 &  -- & -- & -- \\
		& & 1  & 1  & -- & -- & -- \\
		\cline{2-7}
		& \multirow{4}*{\textbf{hGRU}} & \textbf{0} & \textbf{0} & \textbf{0.1887} & \textbf{0.2131} & \textbf{0.2009} \\
		& & 0 & 1  & 0.1739 & 0.2009 & 0.1874 \\
		& & 1  & 0 & 0.3576 & 0.3791 & 0.3684 \\
		& & 1  & 1  & 0.2680 & 0.2841 & 0.2761 \\
		\cline{2-7}
		& \multirow{4}*{LSTM} & 0 & 0 &  0.3690 & 0.3827 & 0.3759 \\
		& & 0 & 1  & 0.3116 & 0.3285 & 0.3201 \\
		& & 1  & 0 & 0.4337 & 0.4583 & 0.4460 \\
		& & 1  & 1  & 0.3567 & 0.3623 & 0.3595 \\
		\hline
		\hline
		\multirow{12}*{4} & \multirow{4}*{Conv} & 0 & 0 & 0.4056 & 0.4173 & 0.4115 \\
		& & 0 & 1  & 0.2896 & 0.3041 & 0.2969 \\
		& & 1  & 0 & -- & -- & -- \\
		& & 1  & 1 & -- & -- & -- \\
		\cline{2-7}
		& \multirow{4}*{hGRU} & 0 & 0 & 0.2347 & 0.2518 & 0.2433 \\
		& & 0 & 1  & 0.2275 & 0.2377 & 0.2326 \\
		& & 1  & 0 & 0.4072 & 0.4284 & 0.4177 \\
		& & 1  & 1 & 0.3080 & 0.3252 & 0.3153 \\
		\cline{2-7}
		& \multirow{4}*{LSTM} & 0 & 0 & 0.4782 & 0.4924 & 0.4853 \\
		& & 0 & 1  &  0.3250 & 0.3441 & 0.3346 \\
		& & 1  & 0 &  0.5963 & 0.6071 & 0.6017 \\
		& & 1  & 1 & 0.3953 & 0.4058 & 0.4006 \\
		\hline
	\end{tabular}
	\caption{Performance for various hyper-parameter configurations. Bold values highlight the best performance.}
	\label{table:prednet_parameters}
	\vspace{-0.7em}
\end{table}
In the following, we discuss the manner in which we exploit information provided by the predictive coding model to estimate the pose of an object.
With camera frames recorded at fixed time intervals $\Delta t_{\text c}$, assume that frame $k \in \mathds{N}$ is available at time instant $t_k = k \Delta t_{\text c}$. We define
\vspace{-0.5em}%
\begin{equation}
x_{\text{o}k} \triangleq x_{\text{o}}(t_k), \quad 
        \Omega_{\text{o}k} \triangleq \Omega_{\text{o}}(t_k),
\end{equation}%
where $x_{\text{o}k}$, $\Omega_{\text{o}k} \in \mathds{R}^3$, describe the object's position (in m) and attitude (in rad) relative to that of the camera at time instant $t_k$, respectively. The attitude is described using Euler angles in the roll-pitch-yaw configuration. We estimate the object's relative pose (position and attitude) by associating segmentation masks with different possible candidate poses, and searching over the range of possible (or likely) poses the one whose associated candidate mask best matches the predicted mask $m_k$, produced from input frame $k$. Candidate mask generation is done using a 3D renderer (we adapted the Godot game engine, \cite{godot2023}). We describe the relationship from candidate pose to candidate mask as follows,
\vspace{-0.35em}%
\begin{equation}
\hat{m}_k = f_{\text g}(\hat{x}_{\text{o}k}, \hat{\Omega}_{\text{o}k}),
\end{equation}%
where $\hat{x}_{\text{o}k}$, $\hat{\Omega}_{\text{o}k}\in \mathds R^3$ describe candidate relative position and attitude, respectively, and $\hat{m}_k \in \mathds{N}^{h \times w}$ is the corresponding candidate mask. The function $f_{\text g}(\cdot)$ represents the rendering function producing masks from object poses (and geometries). To estimate the object's pose at frame $k$, we compare rendered candidate masks $\hat{m}_k$ with the segmentation mask $m_k$ generated by the Predictive Coding model,
\begin{eqnarray}
\left[\begin{array}{c}
    x^*_{\text{o}k} \\
    \Omega^*_{\text{o}k}
\end{array}\right] &\triangleq& 
\argmax_{\hat{x}_{\text{o}k}, \hat{\Omega}_{\text{o}k}} f_{\text m}(m_k, f_{\text g}(\hat{x}_{\text{o}k}, \hat{\Omega}_{\text{o}k})),
\end{eqnarray}%
where $f_{\text m}(m_k,\cdot)$ is a comparison function quantifying the overlap with $m_k$, and $x^*_{\text{o}k}$, $\Omega^*_{\text{o}k}$ denote the candidate relative pose that maximizes this overlap. For ease of exposition, consider a situation in which we segment a single object, such that $m$'s entries are limited to either 1 (if the object is present at the given pixel) or 0 (no object present). Consider the following comparison function,
\begin{equation}
f_{\text m}(m, \hat{m}) = \sum_{i=1}^{h} \sum_{j=1}^{w} \frac{4(m_{ij} - \frac{1}{2}) (\hat{m}_{ij} - \frac{1}{2})}{wh},
\end{equation}%
where $m_{ij}$, $\hat{m}_{ij} \in \{0,1\}$ describe the entries at row $i$, column $j$ of arrays $m$ and $\hat{m}$, respectively. Pixels where $m$ and $\hat{m}$ have equal values (i.e. classification at the current pixel is consistent) provide a positive contribution, whereas inconsistent values lead to negative contributions. Accordingly, the candidate mask $\hat m$ maximizing $\max(f_{\text m}(m, \hat{m}))$ is the candidate most closely matching the prediction mask $m$. In a situation in which we have no prior on the object's pose, we can conduct a search over the entire configuration space for the best mask fit. In a situation in which we do have a prior, assuming that the frame sequence was produced by sampling time-continuous physics, it typically proves of benefit to initialize the search for $x^*_{\text{o}k}$, $\Omega^*_{\text{o}k}$ at the previous instant's $x^*_{\text{o}k-1}$, $\Omega^*_{\text{o}k-1}$. This search is conducted by performing a gradient ascent search on $f_{\text m}(m_k, \cdot)$. As $f_{\text m}(m_k, \cdot)$ describes a non-trivial computational function, obtaining a closed-form expression for the partial derivative of $f_{\rm}(m,\cdot)$ with respect to $\hat x_{\text{o}k}$, $\hat \Omega_{\text{o}k}$ is usually not in practice achievable. Instead, we compute a numerical estimate of the Jacobian using multiple samples around a point of interest in which we want to assess the slope. This numerical Jacobian is used in the aforementioned gradient climb, which allows convergence to a maximum of $f_{\rm}(m,\cdot)$. In the general case of a a search over the entire configuration space, the straightforward application of such a gradient following approach may prove problematic, converging to distant local minima. However, in situations in which prior knowledge is available, and provided non-ambiguous geometry and sufficient frame-rate, we observed that, in practice, the simple gradient following approach performed adequately. Results obtained for the segmentation of a human skeletal forelimb are shown in Fig. \ref{fig:vis_segmentation}, representing frame input (top left), ground truth segmentation (top right), predicted masks (bottom left), and pose estimate (bottom right).  

The visual processing model provides interesting levels of performance; we observed centimetric pose estimation errors at a range of 4m (in rendered benchmarks). Note however that at greater distances, the model struggles in estimating depth with accuracy. Results of a benchmark in which we trained the model to segment and estimate the relative pose of a human skeletal forelimb (as shown in Fig. \ref{fig:vis_segmentation}) are represented in Fig. \ref{fig:vis_sensor_dist_err}. The error in the depth direction (bottom left) rapidly increases at ranges greater than 4m. Instead, position errors in the vertical frame parallel to the camera lens remain moderate.

\begin{table}
	\centering
	\begin{tabular}{|c|c||c||c|}
		\hline %
		\textbf{Camera distance} & \textbf{Level of occlusion} & $e_{\rm v}$ & $e_{\rm nv}$\\
		\hline %
		\hline %
		& \textbf{Average} & \textbf{3.74cm} & 3.92cm \\
		\hline \hline %
		\multirow{3}*{Short (0--3m)} & Light (0--33\%)  & 1.73cm & \textbf{1.62cm} \\
		& Medium (33--66\%) & \textbf{2.48cm} & 4.83cm \\
		& Heavy (66--100\%) & \textbf{5.84cm} & 6.29cm \\
		\hline 
		\multirow{3}*{Medium (3--6m)} & Light (0--33\%) & 1.81cm & \textbf{1.57cm} \\
		& Medium (33--66\%) & \textbf{2.59cm} & 5.02cm \\
		& Heavy (66--100\%) & \textbf{6.04cm} & 6.54cm \\
		\hline 
		\multirow{3}*{Large (6--10m)} & Light (0--33\%) & 4.07cm & \textbf{2.65cm} \\
		& Medium (33--66\%) & 5.39cm & \textbf{2.86cm} \\
		& Heavy (66--100\%) & -- & -- \\
		\hline 
	\end{tabular}
	\caption{ProcNet- ($e_{\rm v}$) and PoseCNN-based pose estimation errors ($e_{\rm nv}$) under various conditions.}
	\label{fig:dataset_results}
\end{table}
	
	\section{Numerical Simulation} \label{sec:numerical_simulation}%
	\begin{figure*}
	\centering
	\includegraphics[width=\textwidth]{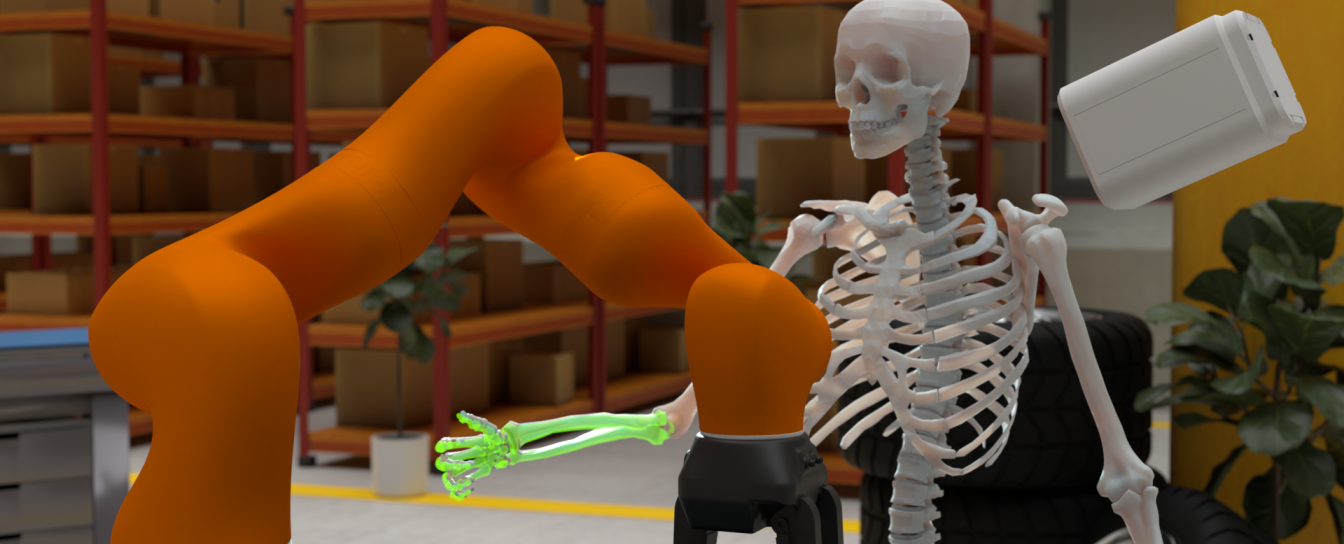}
	\caption{Workspace containing robot arm, human, and camera. The bright green outline shows the estimated pose.}
	\label{fig:scene_setup}
\end{figure*}

\begin{figure}
	\centering
	\includegraphics[width=\columnwidth]{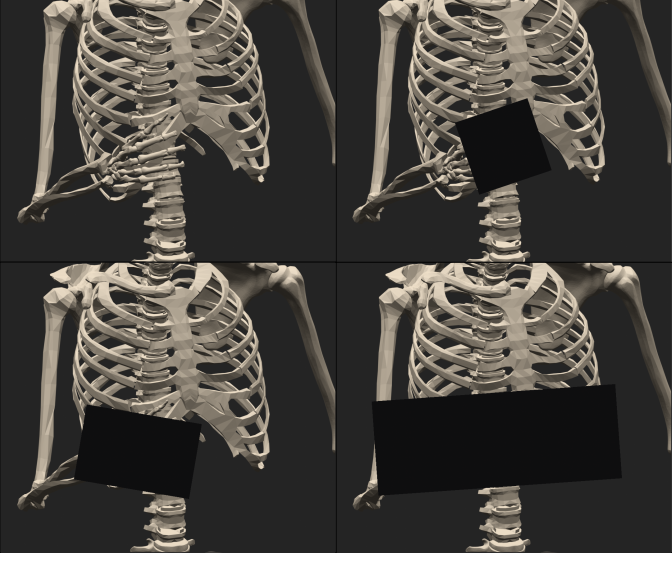}%
	\caption{Example of various levels of occlusion; no occlusion (top left), light occlusion (top right), medium occlusion (bottom left), heavy occlusion (bottom right).}
	\label{fig:occlusion_summary}
\end{figure}

To help assess ProcNet performance, we generated a series of data sets using rendered robotic environments featuring a human (skeletal) model and a robot arm sharing a collaborative work-cell. Both arms are animated, with the robot moving in random trajectories within its six Degrees of Freedom (Dof) workspace, while the human arm is afforded two DoFs in the shoulder, and another in the elbow. The work-cell is set in a warehouse environment with some measure of background clutter (see Fig. \ref{fig:scene_setup}). To quantify performance in relation to controlled levels of occlusion, we artificially obstruct the camera's view by setting a certain amount of image pixels to black (see Fig. \ref{fig:occlusion_summary} for illustration). We quantify occlusion percentage as the ratio of the sum of all relevant but obscured pixels over the sum of all pixels describing a considered object's shape (a frame with a completely visible object features 0\% occlusion, while a frame with a fully obscured object has 100\% occlusion). We divide data sets to distinguish three levels of occlusion: light, medium, and heavy. Similarly, we distinguish frame streams captured at different distances (from camera lens to work-cell center): short, medium, and large (see Table \ref{fig:dataset_results}). We use this data set to investigate the impact of various network hyper-parameters on performance. Then, we investigate ProcNet's performance under various conditions, and compare it to that of an established pose estimation method, NVIDIA's PoseCNN (\cite{xiang2017posecnn}). Datasets are available \href{https://github.com/Mike2208/prednet_localization_results}{here}.
	
	\subsection{Comparison of ProcNet configurations} \label{sec:network_configuration} %
To determine the effects of network hyper-parameters, we trained ProcNet on the same data set in different configurations. In particular, we altered the model's number of layers, representation block layer types, and presence of axonal delay. The results are shown in Table \ref{table:prednet_parameters}; they show that accounting for prediction loss in the backpropagation process reduces Dice loss. Use of hGRU cells in the representation blocks outperforms alternatives. Surprisingly, increasing the amount of layers only improves performance when using convolutional representation layers. In other instances, a ProcNet composed of three layers provides better performance. Similarly, inclusion of axonal delays, which had been speculated to promote emergence of a beneficial temporal integration of information in the psycho-physics literature, negatively impacts performance.
	
	\subsection{Pose Estimation under various levels of occlusion} %
	To investigate performance of ProcNet under different operation conditions, we conducted a series of numerical simulations (described in section \ref{sec:numerical_simulation}). We employed the best ProcNet configuration determined in section \ref{sec:network_configuration} (3 layers, hGRU cells, no axonal delay, and active predictive loss backpropagation). Results are shown in Fig. \ref{fig:dataset_results} and compared to those obtained from the implementation of NVIDIA's PoseCNN (\cite{xiang2017posecnn}). Both approaches failed to produce coherent pose estimates at large distances under heavy occlusion. However, at a range of up to 6m, and for medium to heavy occlusion, ProcNet provides better performance than PoseCNN. Conversely, in instances in which occlusion is limited, PoseCNN performs best.

	
	\section{Conclusion} %
In this paper, we presented ProcNet, a visual segmentation and pose estimation model providing robustness to visual occlusion. The model is developed by extending PredNet using insights from psycho-physics, intended to promote spatial and temporal integration of information. In addition, a decoding stage is included to decode latent information into segmentation masks. The segmentation information is then exploited using a simple generative model to estimate the pose of the considered object. Results of numerical simulations show that, in a rendered robotic work-cell scene, ProcNet is able to provide a measure of robustness to occlusion. Future work will investigate the integration of ProcNet with complementary perception modalities, such as for instance active electric proximity perception, to explore the efficacy of such a multimodal scheme in supporting reliable, robust-to-occlusion human pose estimation in a collaborative robotic setting.
	
	\hypersetup{urlcolor=black}
	\bibliographystyle{IEEEtran}%
	\bibliography{IEEEabrv,references}	
\end{document}